\documentclass[10pt,twocolumn,letterpaper]{article}

\usepackage{iccv}
\usepackage{times}
\usepackage{epsfig}
\usepackage{graphicx}
\usepackage{amsmath}
\usepackage{amssymb}
\usepackage{multirow}
\usepackage{wrapfig}
\usepackage{caption}
\usepackage{enumitem}
\usepackage{booktabs}
\usepackage{flushend}
\usepackage{subcaption}
\usepackage[accsupp]{axessibility}

\usepackage[breaklinks=true,bookmarks=false]{hyperref}

\iccvfinalcopy 

\newcommand{\R}{\mathbb{R}}
\DeclareMathOperator{\dist}{dist}
\DeclareMathOperator{\height}{height}
\DeclareMathOperator{\width}{width}

\def\iccvPaperID{10412} 

\ificcvfinal\fi
\newcommand{\myparagraph}[1]{\smallskip\noindent\textbf{#1}}

\begin{document}

\title{Calibrating Uncertainty for Semi-Supervised Crowd Counting}

\author{Chen Li~\thanks{Email: Chen Li (li.chen.8@stonybrook.edu).} \hskip 1em Xiaoling Hu \hskip 1em Shahira Abousamra  \hskip 1em Chao Chen\\
Stony Brook University\\
\and
}


\maketitle
\ificcvfinal\thispagestyle{empty}\fi

\begin{abstract}
Semi-supervised crowd counting is an important yet challenging task. A popular approach is to iteratively generate pseudo-labels for unlabeled data and add them to the training set. 
The key is to use uncertainty to select reliable pseudo-labels. In this paper, we propose a novel method to calibrate model uncertainty for crowd counting. Our method takes a supervised uncertainty estimation strategy to train the model through a surrogate function. This ensures the uncertainty is well controlled throughout the training. We propose a matching-based patch-wise surrogate function to better approximate uncertainty for crowd counting tasks. The proposed method pays a sufficient amount of attention to details, while maintaining a proper granularity. Altogether our method is able to generate reliable uncertainty estimation, high quality pseudo-labels, and achieve state-of-the-art performance in semi-supervised crowd counting.


\end{abstract}
\section{Introduction}

\begin{figure*}[h]
    \centering 

   \includegraphics[width=1\linewidth]{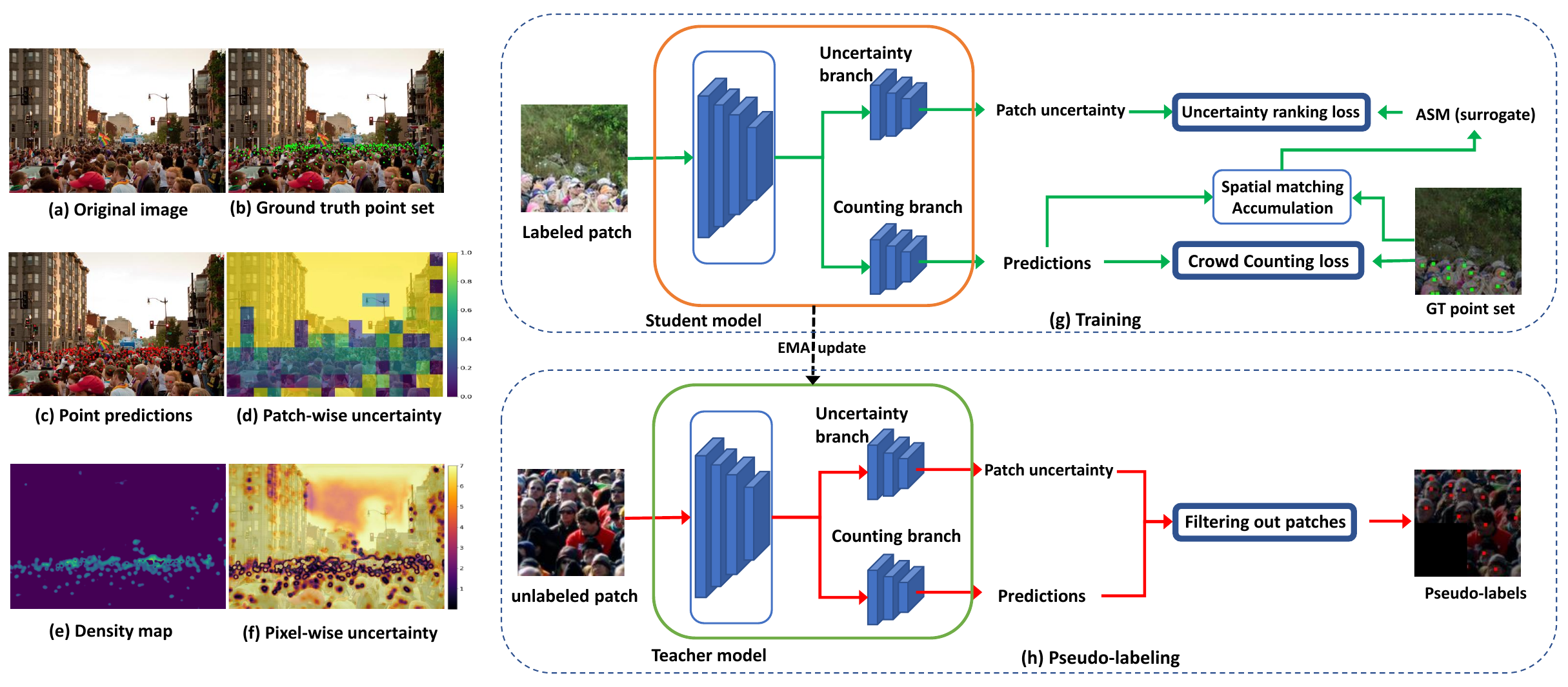}


\caption{An overview of our method. The light color represents low uncertainty in the whole context. \textbf{(a)} Original image. \textbf{(b)} Ground truth point set. \textbf{(c)} Point predictions from our method. \textbf{(d)} Our patch-wise uncertainty map. \textbf{(e)} Density map generated by density-based method~\cite{meng2021spatial}. \textbf{(f)} Pixel-wise uncertainty generated by comparing density maps~\cite{meng2021spatial}. \textbf{(g)} The pipeline of student model training for crowd counting and for uncertainty estimation. A labeled patch goes through the feature extractor and uncertainty branch to generate an uncertainty score. The uncertainty score is compared with the surrogate function (ASM) using uncertainty ranking loss. 
Counting predictions generated by the counting branch are supervised by crowd counting loss.
\textbf{(h)} 
Pseudo label generation process. We use a teacher model which is acquired from the student model via the EMA strategy.
It generates predictions and patch uncertainty for all unlabeled patches. Patches with low uncertainty patches are used to supervise the training of crowd counting. 
}
\vspace{-.1in}
\label{fig:first}
\end{figure*}

Crowd counting is the task of estimating the number of individuals in images or videos.  
It is important for public security, transport management, crowd surveillance, and catastrophe management, to name a few.
Deep-learning-based methods~\cite{wan2019residual, yang2020reverse, zhang2019wide, sindagi2019multi, shi2018crowd, abousamra2021localization, ranjan2018iterative,jiang2019crowd, liu2019attention, shi2019revisiting, shen2018crowd, liu2019context} have achieved promising performance in crowd counting tasks. However, to achieve superior performance, these methods require a large amount of annotations. For each image, hundreds or thousands of points/markers are added to indicate pedestrians (see Fig.~\ref{fig:first}(b)). Acquiring such annotations is costly and time-consuming; it takes 2000 human hours to annotate UCF-QNRF~\cite{idrees2018composition} dataset, which contains 1.25 million points/markers. 

To alleviate the annotation burden, semi-supervised crowd counting is explored to leverage the information in unlabeled images with fewer annotated images. 
Previous works use self-supervised learning~\cite{liu2018leveraging} and data synthesis~\cite{wang2019learning, wang2021pixel}.
Another promising direction is \emph{pseudo-labeling}, i.e., generating pseudo-labels for unlabeled data based on model prediction, and adding them as additional training data. Pseudo-labeling has been used in many other vision tasks, such as  classification~\cite{sohn2020fixmatch}, segmentation~\cite{french2020semi}, and object detection~\cite{xu2021end}. To select reliable predictions as pseudo-labels, one may use the \emph{uncertainty} of the model.
Predictions with low uncertainties are considered more reliable and can be used as pseudo-labels to train the model.

Despite the wide application of uncertainty in computer vision~\cite{moon2020confidence, geifman2018bias, li2023confidence,kohl2018probabilistic, baumgartner2019phiseg, yu2019uncertainty,xu2021end, hu2022learning, gupta2023topology}, estimating uncertainty in crowd counting remains a challenge, due to the heterogeneity of the marker distributions and fundamental issues of a crowded scene including perspective, occlusion, and cluttering.
A few previous works~\cite{meng2021spatial,liu2020semi} estimate uncertainty for crowded scenes through the consistency between predictions made by different models. However, these methods completely rely on models' predictions on unlabeled data. Without a proper controlling mechanism, the uncertainty estimation can not be guaranteed to be reliable. In challenging regions, where all models make similar mistakes, consistency may create overconfidence yet wrong predictions. The training may be derailed by noisy pseudo-labels.

In this paper, we propose a novel semi-supervised crowd counting approach. We focus on uncertainty estimation and propose \emph{the first supervised solution for crowd counting}. Unlike previous methods, we use the labeled data as direct supervision for uncertainty estimation. Although the true uncertainty is unknown even for labeled data, we propose a surrogate function based on the similarity between model predictions and ground truth. Intuitively, when a model can predict well, it should be less uncertain about the data. This learning-performance-based surrogate function provides the opportunity to directly train the model for uncertainty estimation. One appealing feature is that the supervised uncertainty estimation is well under-control throughout different training stages, providing necessary stability to the dynamic semi-supervised learning process.

One critical question is the design of the surrogate function, i.e., how to measure the similarity between a prediction and the ground truth. As the crowd counting prediction can be evaluated in many different ways, a few different similarity measures have been used for uncertainty estimation. Direct evaluation of uncertainty at image level~\cite{zhao2020active} cannot account for the heterogeneous spatial distribution of pedestrians. At the other extreme, Meng et al.~\cite{meng2021spatial} propose pixel-level uncertainty using the density maps, i.e., kernel density based on the points/markers (Fig.~\ref{fig:first}(e)). They compare the density maps of teacher and student models pixel-by-pixel to estimate uncertainty at every pixel. This uncertainty, however, can be unreliable because the density maps lack the necessary details, i.e., exact locations of points/markers. As shown in Fig.~\ref{fig:first}(f), one obtains very low pixel-wise uncertainty at high density regions, where the models tend to be erroneous; but the sky regions get higher uncertainty.

To tackle this challenge, our second contribution is a novel measure of uncertainty that better suits crowd counting tasks.
We focus on the exact coordinates of points/markers and compare the prediction point set and ground truth point set (the set constructed by the coordinates of individuals) through a matching-based similarity metric, called Accumulated Spatial Matching distance (ASM). This matching-based strategy accounts for the full details of the prediction and avoids systematic over-count/under-count bias. To provide the proper granularity and account for the spatial heterogeneity of uncertainty, we partition images into small patches and estimate uncertainty at the patch level. Fig.~\ref{fig:first}(d) illustrates the patch-level uncertainty estimation by our method; the estimated uncertainty is high in crowded regions and low in regions with fewer markers, e.g., sky. This is consistent with our expectations.

We incorporate our supervised uncertainty estimation into an end-to-end semi-supervised learning pipeline. Our model uses a student model and a teacher model. During each training epoch, the student model is trained on the training set for both prediction and uncertainty estimation. The uncertainty training is by comparing with the proposed ASM surrogate function. The teacher model is a stabilized version of the student model; its weights are an exponential moving average (EMA) of the weights of the student model. The teacher model makes predictions and estimates uncertainties on unlabeled patches. Predictions with low uncertainty are chosen to be pseudo-labels and be used to supervise crowd counting.  

In summary, our contribution is three-fold.
\begin{itemize}[topsep=1pt,itemsep=1pt,partopsep=1pt, parsep=1pt]
\item We propose a novel supervised approach to calibrate patch-wise uncertainty for crowd counting task. The estimated uncertainty can be used to effectively select reliable pseudo-labels to enhance semi-supervised training.
\item We propose a novel patch-wise matching-based similarity measure as a surrogate function for uncertainty. This surrogate function focuses on specific coordinates and cardinalities of points/markers and provides reliable information throughout different training stages.
\item On various benchmarks, experimental results show that our method generates well-calibrated uncertainties, high-quality pseudo-labels, and achieves state-of-the-art performance on the semi-supervised crowd counting task.
\end{itemize}


\section{Related Works}

Deep learning based algorithms have achieved great progress in crowd counting tasks. 
In this section, we will review the works in crowd counting, uncertainty estimation, and semi-supervised learning.

\myparagraph{Crowd counting.}
Density-based crowd counting frameworks are widely studied. The crowd counting problem is treated as a density estimation problem by density-based methods. The crowd size is calculated by summing all the pixel values of the density map. 
The density map can be generated pixel-by-pixel~\cite{li2018csrnet, hu2020count, liu2020adaptive, bai2020adaptive, jiang2020attention, miao2020shallow} or patch-by-patch~\cite{xiong2019open, liu2019counting, liu2020weighing}. Wang et al.~\cite{wang2020distribution} utilize Optimal Transport (OT) to measure the distribution difference between the predicted density map and ground truth. Ma et al.~\cite{ma2019bayesian} construct the likelihood function of individuals based on Gaussian distribution. Bai et al.~\cite{bai2020adaptive} address the noise in crowd counting annotations and conduct a self-correction (SC) supervision framework, which can correct annotations based on the model's output. Many promising methods have been published in this track. Despite the strong performance of density-based methods, they cannot generate the accurate locations of individuals out of the predicted density map. Localization-based methods count the crowd by locating all individuals. Some of them utilize object detection techniques to get the individual locations~\cite{sam2020locate, liu2019point, lian2019density}. However, most crowd counting datasets are only point annotated. This makes it difficult to acquire the precise coordinates of bounding boxes and leads to inferior model performance. The locations of individuals are also captured by dots~\cite{liu2019recurrent} or blobs~\cite{laradji2018blobs}, but ad-hoc post-process is used to eliminate false-positive and separate joint individuals. Song et al.~\cite{song2021rethinking} propose a one-to-one matching framework to match the point proposals with ground truth locations. Instead of comparing the location distribution difference, they focus on finding positive proposals and increasing model confidence in these proposals. 


\myparagraph{Counting with limited annotations.} 
It is laborious and time-consuming to annotate images for crowding counting tasks. Semi-/weakly supervised learning methods have been proposed to alleviate the annotation burden. Liu et al.~\cite{liu2018leveraging} introduce a learning-to-rank framework to leverage the unlabeled information contained in the relation between crop size and counting number (image patches should include more individuals than their sub-patches).  Yang et al.~\cite{yang2020weakly} propose a soft-label sorting network to utilize the counting information of crowd numbers rather than the locations of individuals. Xu et al.~\cite{xu2021crowd} propose a density-based framework for partial annotation setting. Sindagi et al~\cite{sindagi2020learning} use an iterative learning framework with Gaussian Process (GP) to leverage unlabeled information and boost model performance. Liu et al.~\cite{liu2020semi} construct a series of surrogate tasks for training a feature extractor on both labeled and unlabeled data with a self-training framework. Liu et al.~\cite{lin2022semi} use a contrastive loss to intensify the learning of density maps on labeled and unlabeled data through density agency. Most previous methods can utilize unlabeled data properly, but the inherent noise in unlabeled supervision can deteriorate the model performance significantly.

\myparagraph{Uncertainty estimation for crowd counting.}
Uncertainty estimation is extensively studied for tasks like segmentation~\cite{kendall2017uncertainties, mehrtash2020confidence, kohl2018probabilistic, baumgartner2019phiseg, li2023confidence} and object detection~\cite{he2019bounding, lu2021geometry}. For crowd counting, uncertainty estimation is important but understudied. 
Meng et al.~\cite{meng2021spatial} estimate pixel-wise uncertainty by comparing density maps generated by teacher and student models. As illustrated in Fig.~\ref{fig:first}(f), the pixel-wise uncertainty can be incorrect, especially in dense areas. Comparing models' output also makes the uncertainty less reliable. This method also treats the problem as a segmentation task and compares segmentation maps between teacher and student models for uncertainty estimation. This additional uncertainty map suffers from the same issues as the density-based uncertainty map.
Zhao et al.~\cite{zhao2020active} introduce uncertainty by comparing the crowd density distribution between different images and claim that the images with crowd dense distribution more dissimilar from labeled images have higher uncertainty. 
Liu et al.~\cite{liu2019exploiting} design a self-supervised proxy task based on the fact that for the same region, the wider range scene should contain more individuals. The number of mistakes made on this proxy task can be seen as an uncertainty estimation.
Our method differs from these previous methods in that it is the first one to use a supervised strategy for uncertainty calibration, ensuring a much more stable uncertainty map during training. Furthermore, our method focuses on the locations and cardinalities of markers and provides patch-level uncertainty maps. This ensures that important details are used while proper granularity is used.


\begin{figure*}[h]
    \centering
    \includegraphics[width=0.8\textwidth]{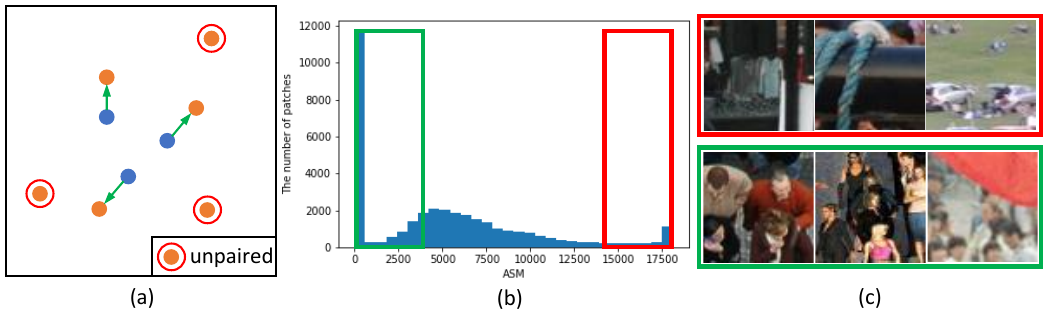}
    \caption{
    \textbf{(a)} An illustration of the introduction of the spatial matching distance. We pair all blue points with orange points one-by-one and use sums of the lengths of the green arrows as part of the matching distance. Three unpaired orange points will contribute a constant penalty each. 
    \textbf{(b)} The ASM distribution on the ShanghaiTech Part-A training set. \textbf{(c)} Sample image patches from ShanghaiTech Part-A, have high ASM (red loop, top) and low ASM (green loop, bottom). The individuals in the red loop are difficult to locate while the crowds in the green loop are easy to count. }
    \vspace{-.1in}
    \label{fig_1}
\end{figure*}

\section{Method}
We propose a pseudo-labeling-based semi-supervised crowd counting approach using patch-level uncertainty estimation. 
Given a small set of labeled images and a large set of unlabeled images, semi-supervised crowd counting is to train models with both labeled and unlabeled images. The key is to make the best use of unlabeled images.
To this end, one effective approach is pseudo-labeling. A pseudo-labeling method assigns to unlabeled data ``pseudo-labels'', which are often model predictions. These pseudo-labeled data will be added to the training set to improve the model. The improved model will then produce better and more pseudo-labels.
The model training and pseudo-label generation can enhance each other in a mutual manner. 

The challenge of the above process is to decide which pseudo-labels are trustworthy. As the model is imperfect, its predictions can be noisy. Picking up too many pseudo-labels will inject noise and derail the training. On the other hand, an overly conservative strategy may not fully utilize the unlabeled data and cannot maximize the potential of the model.
A useful measure for selecting pseudo-labels is uncertainty \cite{sohn2020fixmatch}.
One may choose model predictions with lower uncertainty, hoping that these ``certain'' predictions are more likely to be correct. 

In this paper, we propose a novel uncertainty estimation method for the crowd counting problem and incorporate it into the semi-supervised learning framework. In Sec.~\ref{sec:uncertainty}, we propose to explicitly train uncertainty estimation using labeled data. This way, the uncertainty is more reliable and can be used to select high-quality pseudo-labels. The training is based on a surrogate function that approximates the true uncertainty. We propose a novel surrogate function for crowd counting in Sec.~\ref{sec:asm}. Finally, we introduce the semi-supervised learning framework in Sec.~\ref{sec:semi-supervised}.


\subsection{Uncertainty Calibration Using Labeled Data}
\label{sec:uncertainty}
Our method estimates uncertainty through supervision. This is in contrast with previous semi-supervised crowd counting works~\cite{meng2021spatial,liu2020semi}, which directly calculate uncertainty on unlabeled data based on model prediction consistency. These previous methods, without direct supervision, can make unreliable uncertainty estimations without being noticed, leading to sub-optimal models. 

An overview of our method is illustrated in Fig.~\ref{fig:first}(g). 
Our model takes an image patch as input. It has two branches: one for prediction, and one for uncertainty estimation. 
The two branches share the same feature extractor; this ensures that the uncertainty estimation is based on reliable high-level information and thus generalizes well to unseen data.
During training, our model not only predicts on each unlabeled patch, but also generates uncertainty values for patches. The predictions with low uncertainty will be chosen as pseudo-labels and be added to the training set to refine the model. To ensure the quality of the generated uncertainty, we use labeled data as supervision to train the uncertainty estimation branch. 

The key question is to find an appropriate measure of uncertainty on labeled data. In other words, we need a good ``surrogate function'' as the ``ground truth'' uncertainty.
We draw inspiration from previous work on the classification task. Moon et al.~\cite{moon2020confidence} introduce an uncertainty surrogate function by counting the frequency of correct predictions of snapshot models at different training epochs. The intuition is that a model should be more uncertain on a sample if it is often misclassified. When it comes to the crowd counting task, we need to design a special metric to measure how ``correct'' a prediction is. To this end, we propose a novel similarity measure between ground truth and prediction. We accumulate the similarity across different training epochs. 
Details of the proposed surrogate function, called Accumulated Spatial Matching Distance (ASM), will be presented in Sec.~\ref{sec:asm}.


As illustrated in Fig.~\ref{fig:first}(g), during model training, we not only provide supervision on model prediction, but also supervise the uncertainty branch by comparing its output with the surrogate function. This will ensure that the model can be generalized well to unlabeled data, and produces reliable uncertainty for pseudo-label selection.

\myparagraph{Pairwise ranking loss for uncertainty supervision.} 
To train the uncertainty branch, we propose a pairwise ranking loss. The loss ensures that the rank of uncertainty output is consistent with the rank of uncertainty surrogate. 
Compared with the numeric values of the surrogate function, the ranking is more stable and reliable. This is why ranking loss can achieve better performance than pointwise loss, e.g., L1 loss.
In particular, we use lambdaloss~\cite{wang2018lambdaloss} as follows: 
\begin{equation}
    \mathcal{L}_{uncer} = \sum_{a_i > a_j} |a_i - a_j| \log_2(1 + e^{-(\kappa_i - \kappa_j)}) 
\label{eq:uncertainty-loss}
\end{equation}
where $a_i$ is the batch min-max normalized ASM on patch $I_i$, and $\kappa_i$ is the model's uncertainty output on patch $I_i$. 
The loss will incur a large penalty for a pair of patches with $a_i > a_j$ yet $\kappa_i < \kappa_j$. The loss is also weighted by $|a_i - a_j|$, i.e., the absolute ASM difference between $i$ and $j$.

\subsection{Surrogate for Uncertainty Calibration}
\label{sec:asm}

In this section, we propose a novel measure of similarity between prediction and ground truth for each training patch. Since our focus is to measure the quality of pedestrian point set predictions, we cannot directly use previous similarity measures which compare density maps \cite{wang2020distribution}. We propose match-distance-based metric to compare the predicted and ground truth pedestrian point sets. 
Intuitively, we find an optimal matching between the two point sets, and the matching distance is used as the similarity measure. For each unmatched point, we add a constant penalty. The penalty constant should be proportional to the image patch size, essentially the worst possible matching distance between any two points. 
See Fig.~\ref{fig_1}(a) for an illustration. 

Formally, the proposed similarity between two given pedestrian point sets, \mbox{$P=\{p_i\in \R^2 \mid i=1,\ldots,N_P\}$} and \mbox{$Q=\{q_i\in \R^2 \mid i=1,\ldots,N_Q\}$}. Without loss of generality, we assume $N_P\leq N_Q$. We define the spatial matching distance between them as
\begin{equation}
\dist(P,Q) = \frac{M(P,Q)+H(P,Q)}{N_Q},
\label{eq:spatial-matching}
\end{equation}
in which $M(P,Q)$ and $H(P,Q)$ are the matching distance and the penalty for unmatched points in $Q$. Formally, 
\begin{eqnarray}
M(P,Q) &=& \min\limits_{\gamma \in \Gamma(P,Q)} \sum\nolimits_{p\in P} \| p - \gamma(p) \|_2 \\
H(P,Q) &=& (N_Q-N_P)\cdot C,
\end{eqnarray}
in which $\Gamma$ is the set of all possible one-to-one mappings from $P$ to $Q$.
We choose the penalty $C$ for each unmatched point in $Q$ to be the diagonal length of the image patch, i.e., $C=\sqrt{\height^2+\width^2}$. In practice, for an image patch, $I$, given a prediction pedestrian point set $Pred(I)$ and a ground truth pedestrian point set $GT(I)$, we simply pick the one with the smaller cardinality as $P$, and the other as $Q$. For convenience, we will abuse the notation and denote the distance as $\dist(Pred(I),GT(I))$.


Due to the stochastic nature of deep learning optimizer, the prediction on a single epoch may not be reliable. We stabilize the proposed spatial matching distance by accumulating over snapshot models at different training epochs.
Formally, our accumulated spatial matching distance (ASM) for a training patch, $I$, is as follows:
\begin{equation}
    ASM(I) = \frac{1}{T}\sum\limits_{t = 1}^T \dist(Pred_t(I),GT(I)) 
\end{equation}
where $Pred_t(I)$ is the model predictions at the $t$-th training epoch. $T$ is the total number of training epochs. 
Intuitively, a patch has a lower $ASM$ if it has a better prediction more frequently. A better prediction means the predicted pedestrian point set is better matched with the ground truth points.
As shown in Fig.~\ref{fig_1} (b) and (c), our ASM can be a fairly good surrogate for uncertainty estimation.

\begin{figure}
    \begin{center}
    \includegraphics[width=0.48\textwidth]{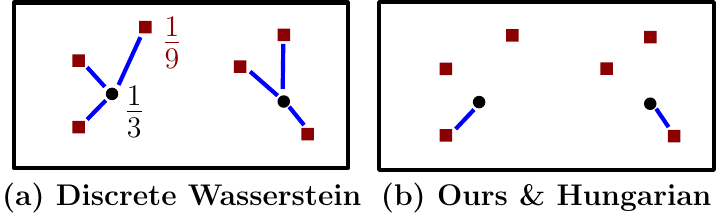}
    \end{center}
        \vspace{-.1in}
    \caption{
     Distance between prediction (red) and GT (black). Discrete Wasserstein gives a fractal mass to each point and matches all of them (blue lines). Our method only conducts one-to-one matching. Unmatched points get a large penalty. Hungarian loss drops unmatched points.}  
    \label{fig_0}
\end{figure}

\myparagraph{Difference between Hungarian loss~\cite{stewart2016end} \& discrete Wasserstein distance.}
The key difference is two-fold. First, our method is the very first to use such loss for the purpose of uncertainty estimation. We have empirically established how powerful it is in semi-supervised crowd counting tasks. Second, from a technical point of view, these methods are not penalizing unmatched points while our method does. The Hungarian loss drops unmatched points as false proposals. The discrete Wasserstein distance treats points as masses and matches them despite a cardinality discrepancy (see Fig.~\ref{fig_0}). Therefore, these methods do not fit our task; we expect a prediction with significantly fewer/more points than GT to have a large uncertainty/penalty. 

\begin{figure*}[h]
    \centering
    \includegraphics[width=1\textwidth]{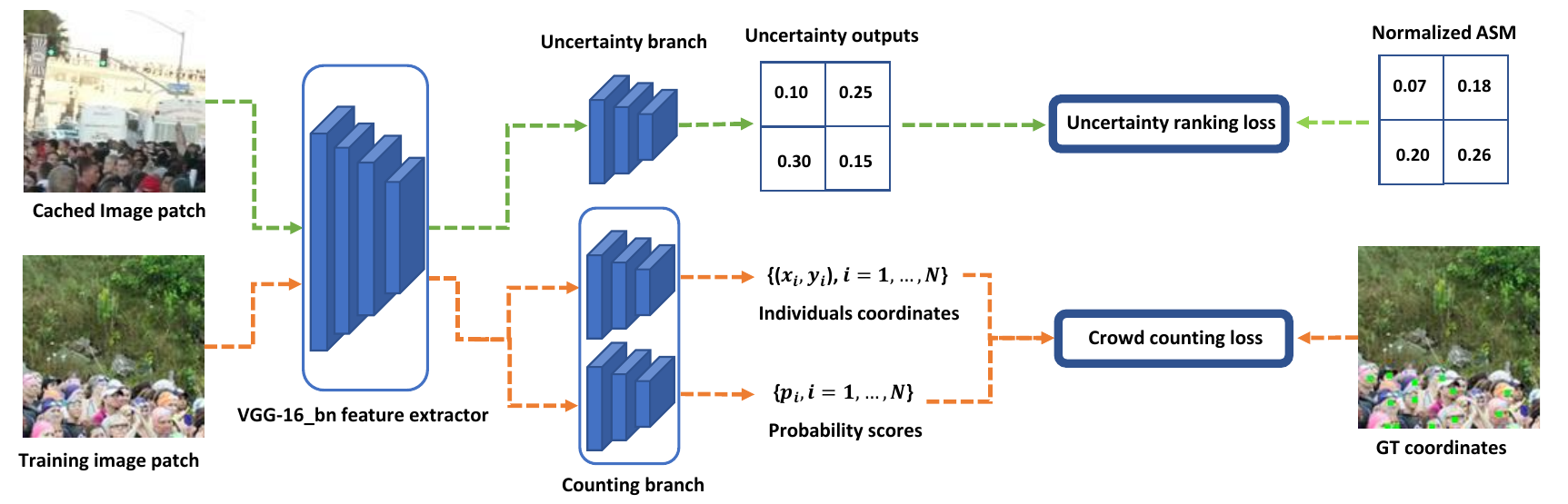}
    \caption{The pipeline of model training for crowd counting and uncertainty estimation. The cached image goes through the feature extractor and uncertainty branch for generating uncertainty scores. These uncertainty scores are supervised by ASM using uncertainty ranking loss. The training image patches are cropped from training images at random locations, and their extracted features are fed into the counting branch for generating the coordinates and probability scores of individuals, which are supervised by ground truth through crowd counting loss. }
    \vspace{-.1in}
    \label{fig_uncer}
\end{figure*}

\myparagraph{Why patch uncertainty is better than pixel uncertainty?}
Pixel-wise uncertainty only uses local information. It is similar to an image segmentation uncertainty map, in which only pixels near object boundaries have high uncertainty. Similarly, as illustrated in Fig.~\ref{fig_qua} (Pixel Uncertainty column), the pixel-wise uncertainty map has high uncertainty between dense and sparse regions. But inside a dense region or inside a sparse region, the uncertainty is low despite the prediction quality. On the other hand, our method uses a patch-wise matching-based surrogate function for uncertainty. It is reliable even inside dense or sparse regions. 

\myparagraph{Implementation details: patch bank.} It is a common practice in crowd counting to extract patches at random locations during training. These random patches are different across different training epochs. Thus we cannot accumulate their matching distances. To address this issue, we choose a fixed patch extraction strategy. We cut input images into a fixed set of patches, and store all these patches into a patch bank. During training, for every patch in the bank, we accumulate their matching distance as a surrogate measure for uncertainty. These patches are randomly drawn to train the model within each training batch. As shown in Fig.~\ref{fig_uncer}, a cached image patch with size $128\times128$ is fed into a model for generating 4 uncertainty outputs. Each output represents the uncertainty of the corresponding $64\times64$ patch region. The uncertainty outputs are supervised by normalized ASM through uncertainty ranking loss. Training image patches, cropped from training images randomly, are used to train the model for crowd counting.

\newcommand{\calD}{\mathcal{D}}
\newcommand{\calL}{\mathcal{L}}
\subsection{Semi-Supervised Crowd Counting}
\label{sec:semi-supervised}
During training, our method starts with a labeled data set $\calD_L$ and an unlabeled data set $\calD_U$. As the training continues, unlabeled data with low uncertainty are assigned pseudo-labels, constituting a pseudo-labeled data set, $\calD_{Pseudo}$. 

We use two models: a teacher model, $f_{tea}$, and a student model, $f_{stu}$. The teacher model makes predictions on unlabeled data set. It generates predictions and estimates uncertainties for all unlabeled patches. Predictions with low uncertainty are chosen as pseudo-labels. These labels, together with their corresponding patches, are added to $\calD_{Pseudo}$. Both $\calD_L$ and $\calD_{Pseudo}$ are used for training. Meanwhile, the student model continuously learns from the training set to make predictions and estimate uncertainty. To train the student model to predict, we use a standard crowd counting loss, $\calL_{pred}$, which is backbone dependent. To train the student model to estimate uncertainty, we use the uncertainty loss $\calL_{uncer}$, as defined in Eq.~\ref{eq:uncertainty-loss}. Altogether, the student model is trained with the loss
\begin{multline}
\calL(f_{stu},\calD) = \calL_{pred}(f_{stu}, \calD_L) + \calL_{uncer}(f_{stu}, \calD_L) + 
 \\ \lambda_1\calL_{pred}(f_{stu}, \calD_{Pseudo}) ,
\label{eq:total-loss}
\end{multline}
in which the weight $\lambda_1$ is tuned empirically.
Note that while both labeled and pseudo-labeled data are used to train prediction, we only use the original labeled data to train uncertainty, in order to ensure a reliable uncertainty estimation. 


We use the same backbone for both the teacher and the student models. In particular, we use P2PNet \cite{song2021rethinking} and its associated prediction loss, $\calL_{pred}$. 
Both models are initialized with the weights pre-trained on ImageNet. We follow the common practice in semi-supervised learning~\cite{sohn2020fixmatch, tarvainen2017mean} and update the teacher model based on the student model through the exponential mean average (EMA) strategy. With this strategy, the teacher network is more stable and reliable. We depend on its prediction and uncertainty estimation to generate pseudo-labels.

We conclude this subsection by explaining our strategy of pseudo-label generation. At every training epoch, we apply the teacher model to all unlabeled data, generating predictions and uncertainties. For each data (image patch), we compare its uncertainty with a preset uncertainty threshold, $u_t$. We select all patches with $<u_t$ uncertainty and their predictions as the pseudo-labeled set. See Fig.~\ref{fig:first} for illustrations. 
Empirically, we adjust the threshold $u_t$ as training progresses. We start with a warm up period, during which we only train the student model with labeled patches. After the warm up period, we use a linearly increasing threshold $u_t$ to select pseudo-labeled patches and add them to the training set. This ensures the pseudo-label generation to be conservative in the beginning and more aggressive later, when the model is more reliable. When adding pseudo-labeled patches, we also apply strong augmentation, i.e., cutout. More details are provided in the supplementary.

\myparagraph{Model architecture details.}
We use the first 13 convolution layers of model VGG-16\_bn~\cite{simonyan2014very} as the features extractor. The features are fed into two branches: the counting branch is the same as in \cite{song2021rethinking}. The uncertainty branch comprises of convolution layers with ReLU activations. The final layer uses a sigmoid activation to generate the uncertainty.

\section{Experiments}
In this paper, we conduct extensive experiments on five public datasets to evaluate the effectiveness of our method:  ShanghaiTech Part-A
and Part-B~\cite{zhang2016single}, UCF-QNRF~\cite{idrees2018composition}, NWPU-Crowd~\cite{wang2020nwpu}, JHU-Crowd++~\cite{sindagi2019pushing, sindagi2020jhu-crowd++}. More implementation details can be found in Appendix. 
\setlength{\tabcolsep}{2.5pt}
\begin{table}[h]
\begin{center}
\begin{tabular}{c|c|cc|cc} 
\hline
\multirow{2}{4em}{Method} & \multirow{2}{2em}{Ratio} & \multicolumn{2}{|c|}{Part A} & \multicolumn{2}{c}{Part B}  \\
                           &                         & MAE&RMSE&MAE&RMSE\\
\hline
SUA~\cite{meng2021spatial} & 50\% & 68.5 & 121.9 & 14.1 & 20.6 \\
GP~\cite{sindagi2020learning} & 25\% & 91& 149 & - & -\\
\hline
MT~\cite{tarvainen2017mean} & 10\% & 94.5 & 156.1 & 15.6 & 24.5 \\
L2R~\cite{liu2018leveraging} & 10\% & 90.3 & 153.5 & 15.6 & 24.4 \\
IRAST~\cite{liu2020semi} &  10\% & 86.9 & 148.9 & 14.7 & 22.9 \\ 
IRAST+SPN~\cite{liu2020semi} & 10\% & 83.9 & 140.1 & - & - \\
AL-AC~\cite{zhao2020active} & 10\% & 80.4 &  138.8 & 12.7 & 20.4 \\
PA~\cite{xu2021crowd} & 10\% & \underline{72.79} & \textbf{111.61} &  12.03 & \underline{18.70} \\ 
DAcount~\cite{lin2022semi} & 10\% & 74.9 & \underline{115.5} & \underline{11.1} & 19.1 \\ 
Ours & 10\% & \bf 70.76 & 116.62 & \bf 9.71 & \bf 17.74 \\
\hline

\end{tabular}
\caption{Results on the ShanghaiTech dataset.}\label{SAB}
\end{center}
\vspace{-.05in}
\end{table}


\myparagraph{Data processing.}
During training, we randomly scale input images with scaling range $[0.7, 1.3]$ and crop patches of size $128 \times 128$. The cropped patches are randomly flipped with a probability of 0.5. For some datasets with very high resolution images, e.g., QNRF-Crowd, JHU-Crowd++ and NWPU-Crowd, we rescale the images so the max sizes of images are shorter than a certain length. Following the settings in P2PNet~\cite{song2021rethinking}, for QNRF-Crowd, JHU-Crowd++, and NWPU-Crowd, this length is 1408, 1430, and 1920.

\myparagraph{Evaluation metrics.}
We use two very common metrics, Mean Absolute Error (MAE) and Root Mean Squared Error (RMSE), to evaluate the model performance.

\myparagraph{Hyperparameters.}
Here we estimate the uncertainty for $64 \times 64$ image patches. Thus the penalty constant is $C = 64\sqrt{2}$. Our method is trained by Adam~\cite{kingma2014adam} with a mini-batch size of 8 using a learning rate of 1e-5 for the parameters of the feature extractor and 1e-4 for the rest model parameters. The weight $\lambda_1$ is set as 0.3. The uncertainty surrogates for images in the bank are updated every training cycle on unlabeled images. 

\myparagraph{Baselines.}
To show how our proposed semi-supervised approach can better utilize unlabeled images and boost performance on crowd counting, we compare its performance against SOTA methods from three tracks: semi-supervised learning (SSL), active learning (ACL), and partial-supervised learning (PAL). 


\subsection{Results}
In this part, we compare our semi-supervised method, trained with only $10\%$ of the ground truth labels, to various baselines and show the superiority of our method through evaluation metrics MAE and RMSE. More experiment results can be found in Appendix.

\begin{table*}[h]
  \begin{minipage}{0.32\textwidth}
  \footnotesize
    \centering
    \scriptsize
\begin{tabular}{ c|cc|cc } 
\hline
Method & Type &Ratio & MAE & RMSE \\
\hline
SUA~\cite{meng2021spatial} & SSL & 50\% & 130.3&  226.3 \\
GP~\cite{sindagi2020learning} & SSL & 25\% & 147 & 226 \\
IRAST~\cite{liu2020semi} & SSL & 20\% & 135.6 & 233.4 \\ 
\hline
MT~\cite{tarvainen2017mean} &  SSL & 10\% & 145.5 & 250.3 \\
L2R~\cite{liu2018leveraging} & SSL & 10\% & 148.9 & 249.8 \\
PA~\cite{xu2021crowd} & PAL & 10\% & 128.13 & 218.05 \\ 
DACount~\cite{lin2022semi} & SSL & 10\% & \underline{109.0} & \underline{187.2} \\
Ours & SSL & 10\% & \textbf{104.04} & \textbf{164.25} \\
\hline

\end{tabular}
\caption{Results on the UCF-QNRF.}\label{QNRF}
  \end{minipage}
  \hfill
  \begin{minipage}{0.32\textwidth}
  \footnotesize
    \centering
    \scriptsize
\begin{tabular}{ c|cc|cc } 
\hline

Method & Type &Ratio & MAE & RMSE \\
\hline
SUA~\cite{meng2021spatial} & SSL & 50\% & 80.7&   290.8 \\
\hline
MT~\cite{tarvainen2017mean} &  SSL & 10\% & 90.2 & 319.3 \\
L2R~\cite{liu2018leveraging} & SSL & 10\% & 87.5 & 315.3 \\
PA~\cite{xu2021crowd} & PAL & 10\% & 129.65 & 400.47 \\ 
DACount~\cite{lin2022semi} & SSL & 10\% & \underline{75.9} & \underline{282.3} \\
Ours & SSL & 10\% & \textbf{74.87} & \textbf{281.69} \\
\hline

\end{tabular}
\caption{Results on JHU-Crowd++.}\label{JHU}
  \end{minipage}
  \hfill
 \begin{minipage}{0.32\textwidth}
  \footnotesize
		\centering
  \scriptsize
\begin{tabular}{ c|cc|cc } 
\hline
Method & Type &Ratio & MAE & RMSE \\
\hline
MT~\cite{tarvainen2017mean} & SSL & 50\% & 129.8 & 515.0 \\
L2R~\cite{liu2018leveraging} & SSL & 50\% & 125.0 & 501.9 \\
SUA~\cite{meng2021spatial} & SSL & 50\% & 111.7&   443.2 \\
\hline
PA~\cite{xu2021crowd} & PA & 10\%  &178.70 &  1080.43 \\
Ours & SSL & 10\% & \textbf{108.78} & \textbf{458.02} \\
\hline

\end{tabular}
\caption{Results on NWPU-Crowd.}\label{NWPU}
  \end{minipage}
\end{table*}

\begin{figure*}[h]
    \centering
    \includegraphics[width=1\textwidth]{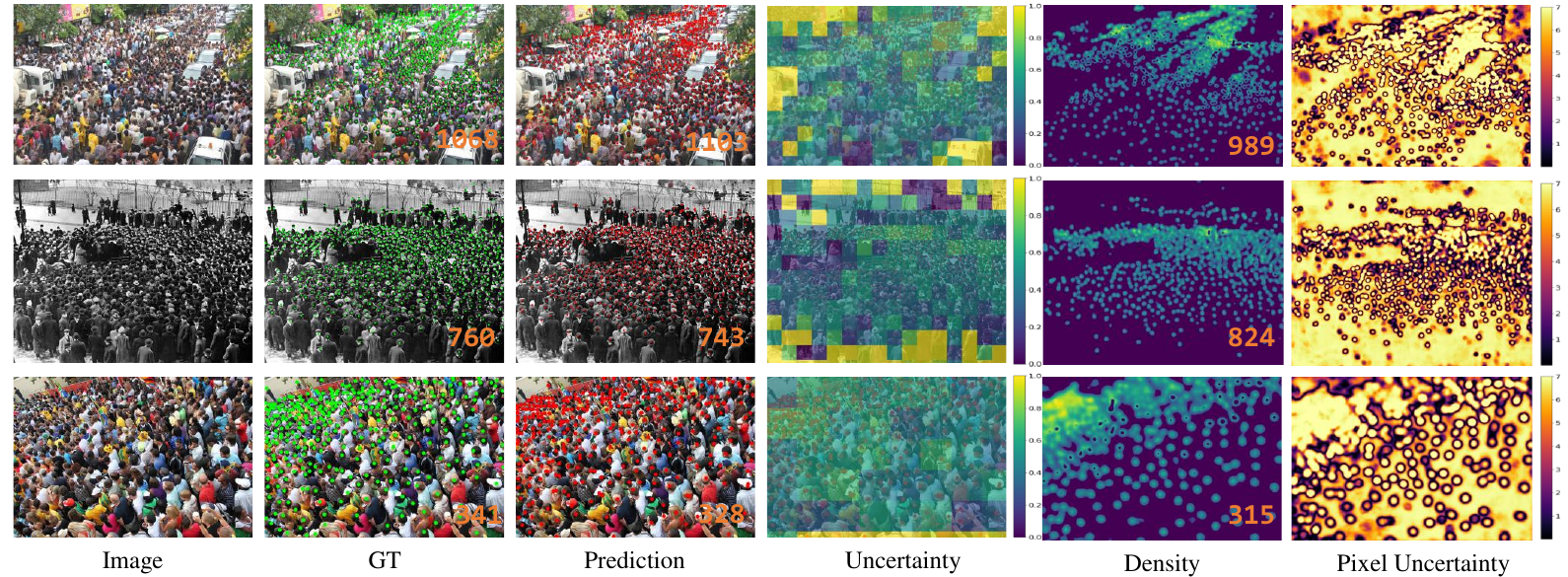}
    \caption{Qualitative results on ShanghaiTech Part-A. The "Prediction" column shows our prediction results. The "Uncertainty" column shows our patch-wise uncertainty, in which yellow represents low uncertainty. The "Density" column is the density map generated by Meng et al.~\cite{meng2021spatial} and the "Pixel uncertainty" column shows their pixel uncertainty.  }
    \label{fig_qua}
\end{figure*}

\myparagraph{ShanghaiTech.}
The ShanghaiTech dataset is constructed by two subsets: Part-A and Part-B. The images of Part-A contain dense crowds collected from the Internet. The Part-B images are acquired from a street in Shanghai, which has relatively sparse crowds. Our method achieves superior performance on both Part-A and Part-B, especially in the MAE metric. Tab.~\ref{SAB} indicates that our method can handle both congested and sparse crowds properly and utilize the information from unlabeled images in a good manner.

\myparagraph{UCF-QNRF.}
UCF-QNRF is a challenging crowd counting dataset due to the diversity of viewpoints, crowd densities, and lighting conditions. Our method has a very good performance on this dataset.  The results in Tab.~\ref{QNRF} indicate that our method is robust to the distribution of the complex background and can generate reliable pseudo-labels under multiple scene situations,  achieving lower MAE and RMSE over SOTA semi-supervised methods.

\myparagraph{JHU-Crowd++.}
JHU-Crowd++ is a large-scale crowd counting dataset containing images with weather-based degradations and multiple environmental conditions. Our method achieves better results than all baselines on both MAE and RMSE with DACount~\cite{lin2022semi} at a close second. The results shown in Tab.~\ref{JHU} reflect that our method can utilize the crowd information in unlabeled images efficiently under various scenarios. 

\myparagraph{NWPU-Crowd.}
NWPU-Crowd is a massive crowd counting dataset containing images with highly congested crowds and large appearance variations. As shown in Tab.~\ref{NWPU}, our method has the best results among weakly supervised algorithms. Besides, our method reduces MAE by 39.1\% and RMSE by 57.6\% compared with the partial annotation-based method (PA)~\cite{xu2021crowd}.

The results show that our semi-supervised method consistently outperforms the state-of-the-art semi-supervised, partial-annotation, and active learning methods. This reflects that our uncertainty estimation is efficient for filtering out low-quality pseudo-labels. Hence, our semi-supervised framework can utilize the information in unlabeled images more effectively. Sample results are in Fig.~\ref{fig_qua}.

\subsection{Ablation Studies}
We conduct experiments to illustrate the effectiveness of each component in our method and the effect of changing hyperparameters and experiment settings.

\myparagraph{The proportion of labeled images.}
In the previous section, we conduct experiments with 10\% labeled images on ShanghaiTech Part-A. To show the effect of labeled image ratio on the performance of our method, here we use extra experiment results on 5\% and 40\% labeled images to validate our method. As shown in Tab.~\ref{rat}, our method achieve better performance than baselines despite the ratio of the labeled images. These results indicate our method can utilize unlabeled images efficiently with different proportions of labeled images. 

\begin{table}[h]
\begin{center}
\begin{tabular}{ c|cc|cc } 
\hline
\multirow{2}{4em}{Method} & \multirow{2}{2em}{Type} & \multirow{2}{2em}{Ratio} & \multicolumn{2}{|c}{Part A}  \\
                           &        &              &MAE&RMSE\\
\hline
MT~\cite{tarvainen2017mean} & SSL& 5\% & 104.7 & 156.9  \\
L2R~\cite{liu2018leveraging} & SSL& 5\% & 103.0 & 155.4 \\
GP & SSL & 5\% & 102.0 & 172.0 \\
PA~\cite{xu2021crowd} & PAL &5\% & \underline{79.42} &  \bf 123.60  \\
DAcount~\cite{lin2022semi} & SSL & 5\% & 85.4 & 134.5  \\
ours &SSL & 5\% & \bf 74.48 & \underline{127.51}  \\
\hline
MT~\cite{tarvainen2017mean} &SSL & 40\% & 88.2 & 151.1 \\
L2R~\cite{liu2018leveraging} & SSL & 40\% & 86.5 & 148.2 \\
DAcount~\cite{lin2022semi} &SSL & 40\% & \underline{67.5} & \underline{110.7}  \\
Ours &SSL & 40\% & \bf 64.74 & \bf 109.56  \\
\hline

\end{tabular}
\caption{The ablation study results of labeled ratio on the ShanghaiTech dataset.}
\label{rat}
\end{center}
\vspace{-.05in}
\end{table}
\myparagraph{Components.}
Here we use experiments on 10\% ShanghaiTech Part-A to verify the effectiveness of components for our method. The results are shown in Tab.~\ref{acom}. Sup.only is the result of the model trained only with supervised counting loss. The result using supervised counting loss and uncertainty loss is shown in Sup.+uncer. There is little improvement in counting performance using our uncertainty loss. To verify the effectiveness of Lambdaloss, here we use L1 loss to supervise the learning process of uncertainty estimation. The results indicate the uncertainty estimation learned with Lambdaloss is more reliable and thus achieves better results. Besides, we also conduct experiments for strong augmentation and EMA. As shown in W/o strong, the lack of strong augmentation has a negative effect on the model performance. The results of W/o EMA indicate EMA is important for utilizing pseudo-labels during the training. An ablation study shows that Hungarian loss has worse performance than our spatial matching distance (Hungarian in Tab.~\ref{acom}).


\begin{table}[ht]
  \begin{minipage}{0.22\textwidth}
  \footnotesize
    \centering
    \scriptsize
\begin{tabular}{ c|c|cc } 
\hline
\multirow{2}{4em}{Method} & \multirow{2}{2em}{Ratio} & \multicolumn{2}{|c}{Part A}  \\
                           &                         & MAE&RMSE \\
\hline
W/o filtering & 10\% & 83.28 & 172.97  \\
Softmax & 10\% & 75.47 & 129.35 \\
ACD & 10\% & 71.90 & 122.03\\
AWD & 10\% & 72.47 & 126.12 \\
w/o average & 10\% & 71.47 &120.07 \\ 
Ours & 10\% & \textbf{70.76} & \textbf{116.62} \\
\hline

\end{tabular}
\caption{The ablation study results of Uncertainty estimation on the ShanghaiTech Part-A.}\label{aun}
  \end{minipage}
  \hfill
  \begin{minipage}{0.22\textwidth}
  \footnotesize
    \centering
    \scriptsize
\begin{tabular}{ c|c|cc } 
\hline
\multirow{2}{4em}{Method} & \multirow{2}{2em}{Ratio} & \multicolumn{2}{|c}{Part A}  \\
                           &                         & MAE&RMSE \\
\hline
Sup.only & 10\% & 77.74 & 125.86  \\
Sup.+uncer. & 10\% & 74.84 & 122.95 \\
L1 loss & 10\% & 72.60 & 117.72\\
W/o strong & 10\% & 72.77 & 124.64 \\
W/o EMA & 10\% & 75.38 & 124.48 \\
Hungarian & 10\% & 72.95 & 121.24 \\
ours & 10\% & \textbf{70.76} & \textbf{116.62} \\

\hline

\end{tabular}
\caption{The ablation study results of components on the ShanghaiTech Part-A.}\label{acom}
  \end{minipage}
  \hfill
\end{table}

\myparagraph{Uncertainty estimation.}
We study the effect of uncertainty estimation on our semi-supervised method. Here we show the superiority of our uncertainty estimation method through experiments on 10\% ShanghaiTech Part-A. We have three baselines for uncertainty estimation: w/o filtering, softmax, accumulated counting difference (ACD), and accumulated Wasserstein distance (AWD). W/o filtering here represents the baseline without filtering out high uncertainty patches. Softmax is the patch uncertainty estimation calculated by the mean confidence scores of the point proposals in each image patch. ACD is the uncertainty surrogate defined by substituting our spatial matching distance with the absolute counting difference between ground truth and prediction. AWD uses discrete Wasserstein distance instead of spatial matching distance to measure the localization difference. To deal with the case when prediction (ground truth) counting is zero and ground truth (prediction) counting is non-zero for AWD, we apply punishment on such case by constructing a super-pixel, the distance of which to all ground truth points is penalty constant C. As shown in Tab.~\ref{aun}, our method achieves the best performance among all baselines, which reflects that our uncertainty estimation is reliable for choosing high-quality pseudo-labels. From the result of W/o filtering, we know the noisy pseudo-labels are detrimental to the training process and can lead to inferior results. Due to the severe overconfidence problem, the pseudo-labels generated with softmax are still noisy. ACD ignores the location information of individuals in crowds, which is critical for estimating model uncertainty. As discussed in Sec.~\ref{sec:asm}, limited by the drawback of discrete Wasserstein distance in evaluating point distribution difference for crowd counting, the performance of AWD is thus worse than our method. In Tab.~\ref{acom} w/o average, we show the necessity of accumulating sptial matching distance during training.

\myparagraph{Hyperparameters.}
In Tab.~\ref{ahyp}, we study the effect of the weight $\lambda_1$ in Eq.~\ref{eq:total-loss}, the maximum value of uncertainty threshold $u_t$, and patch size for uncertainty estimation. We can see our method is not sensitive to those hyperparameters. Our method achieves fairly good performance with the perturbation of hyperparameters.

\begin{table}[h]
\begin{center}
\begin{tabular}{ ccc|c|cc } 
\hline
Patch size & $\lambda_1$ & max $u_t$ & Ratio  & MAE&RMSE \\
\hline
8 & 0.3 & 0.6 & 10\% &76.31 & 126.19 \\
16 & 0.3 & 0.6 & 10\% &77.47 & 133.69 \\
32 & 0.3 & 0.6 & 10\% & 71.37 & 116.74  \\
\bf 64 & \bf 0.3 & \bf 0.6 &  10\% & \bf 70.76 & \bf 116.62 \\
128 & 0.3 & 0.6 & 10\% & 75.41 & 119.94\\
\hline
64 & 0.1 & 0.6 & 10\% & 72.73 & 120.18  \\
64 & 0.2 & 0.6 & 10\% & 71.79 & 119.50\\
\bf 64 & \bf 0.3 & \bf 0.6 & 10\% & \bf 70.76 & \bf 116.62 \\
64 & 0.4 & 0.6 & 10\% & 72.09 & 124.64 \\
64 & 0.5 & 0.6 & 10\% & 70.98 & 122.00 \\
\hline
64 & 0.3 & 0.8 & 10\% & 73.15 & 127.48\\
64 & 0.3 & 0.7 & 10\% & 71.82 & 120.23\\
\bf 64 & \bf 0.3 & \bf 0.6 & 10\% & \bf 70.76 & \bf 116.62\\
64 & 0.3 & 0.5 & 10\% & 73.11 & 122.18 \\
64 & 0.3 & 0.4 & 10\% & 72.87 & 124.39 \\
\hline

\end{tabular}
\caption{The ablation study results of hyperparameters on the ShanghaiTech Part-A.}\label{ahyp}
\end{center}
\end{table}



\vspace{-.2in}
\section{Conclusion}
In this work, we introduce a novel patch-wise uncertainty estimation for pseudo-labeling-based semi-supervised crowd counting. Our method trains the uncertainty estimator directly through a surrogate function calculated on labeled patches. As for the uncertainty surrogate function, we use a spatial matching distance between predictions and ground truth labels. Our method provides reliable uncertainty estimation, thus helping to select pseudo-labels to improve the model training in a semi-supervised fashion. 
The evaluation of our proposed semi-supervised method on several popular crowd counting benchmarks shows that our method consistently achieves superior performance compared to SOTA semi-supervised methods.

\myparagraph{Acknowledgements.} 
We thank the anonymous reviewers for their constructive feedback. 
This work was supported by the NSF grant CCF-2144901.

{\small
\bibliographystyle{ieee_fullname}
\bibliography{egbib}
}

\newpage
\appendix
\section{Appendix}



Append.~\ref{quality_filtering_out} shows quality results about which high uncertain patches are filtered out.

Append.~\ref{is5ok} shows that our method is reliable even with only 5\% labeled samples.

Append.~\ref{moreab} provides more ablation study results.



Append.~\ref{sec:peudo-label} illustrates the implementation details of choosing pseudo-labels.

\subsection{Illustrate which pseudo-labels were selected and which inaccurate ones were filtered out.} 
\label{quality_filtering_out}

\begin{figure}[h]
\centering 
  \begin{subfigure}{0.24\linewidth}
   \includegraphics[width=1\linewidth]{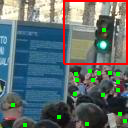}
  \end{subfigure}
  \begin{subfigure}{0.24\linewidth}
   \includegraphics[width=1\linewidth]{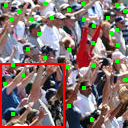}
  \end{subfigure}
  \begin{subfigure}{0.24\linewidth}
   \includegraphics[width=1\linewidth]{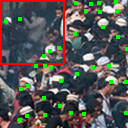}
  \end{subfigure}
  \caption{Qualitative results on uncertainty filtering.}
  \label{quau}
\end{figure}

Fig.~\ref{quau} shows results on unlabeled samples. Green dots are predictions and red rectangles are high-uncertainty patches. On the left, we found a high-uncertainty patch within the sparse region, containing only one false positive (on the traffic light). In the middle and right samples, the high-uncertainty patches contain many false negatives due to occlusion or dark shades.


\subsection{Is 5\% labeled data enough for training reliable
uncertainty estimator?}
\label{is5ok}

Empirical results show that 5\% labeled data is sufficient to achieve superior performance on ShanghaiTech A (main paper Tab.~\ref{rat} \& Tab.~\ref{aun_B}) and B (Tab.~\ref{rat_B}) datasets. 

\setlength{\tabcolsep}{2pt}
\begin{table}[h]
  \footnotesize
    \centering
    \scriptsize
\begin{tabular}{ c|c|cc } 
\hline
\multirow{2}{4em}{Method} & \multirow{2}{2em}{Ratio} & \multicolumn{2}{|c}{Part A}  \\
                           &                         & MAE&RMSE \\
\hline
sup.only & 5\% & 88.48 &162.42 \\
w/o filtering & 5\% & 111.39 & 174.87 \\
 softmax & 5\% & 94.64 & 155.65 \\
 Ours & 5\% & \bf 74.48  &\bf 127.51 \\ 
\hline

\end{tabular}
\vspace{-.1in}
\caption{The ablation study results of 5\% labeled data.}\label{aun_B}
  \hfill
  \vspace{-.15in}
\end{table}

\subsection{Extra ablation study results.}
\label{moreab}


In this section, we show the effectiveness of our method under 5\% and 40\% labeled images on the ShanghaiTech part-B dataset with extra ablation study experiments. As shown in Tab.~\ref{rat_B}, our method achieves better performance under both 5\% and 40\% labeled image scenarios. This indicates our method can obtain superior performance for semi-supervised crowd counting under various labeled ratios on different datasets.

\begin{figure}[h]
\centering 
  \begin{subfigure}{0.34\linewidth}
   \includegraphics[width=1\linewidth]{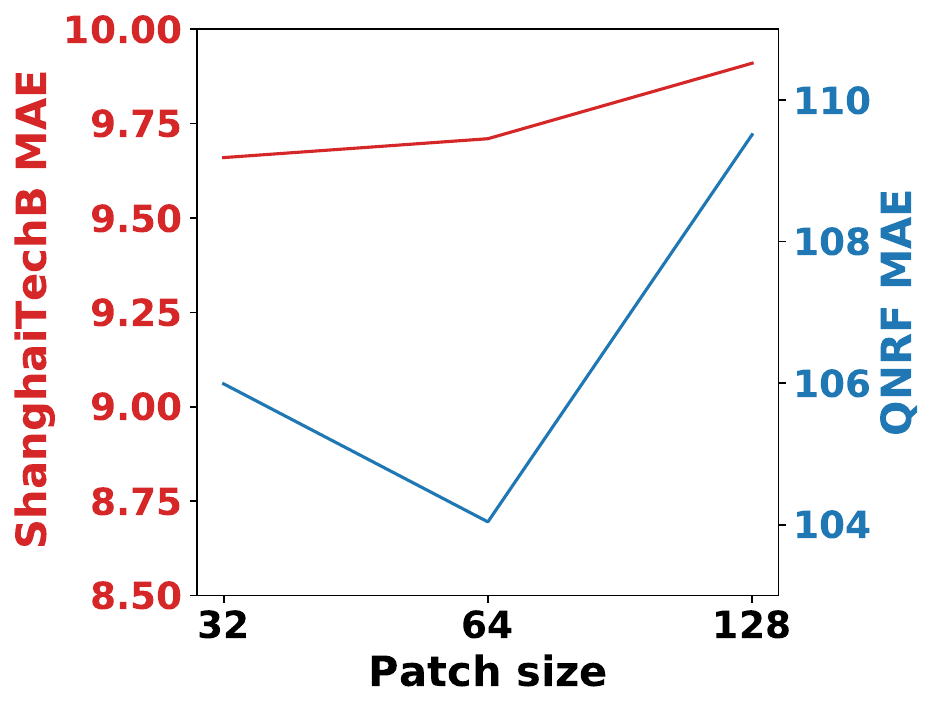}
  \end{subfigure}
  \begin{subfigure}{0.31\linewidth}
   \includegraphics[width=1\linewidth]{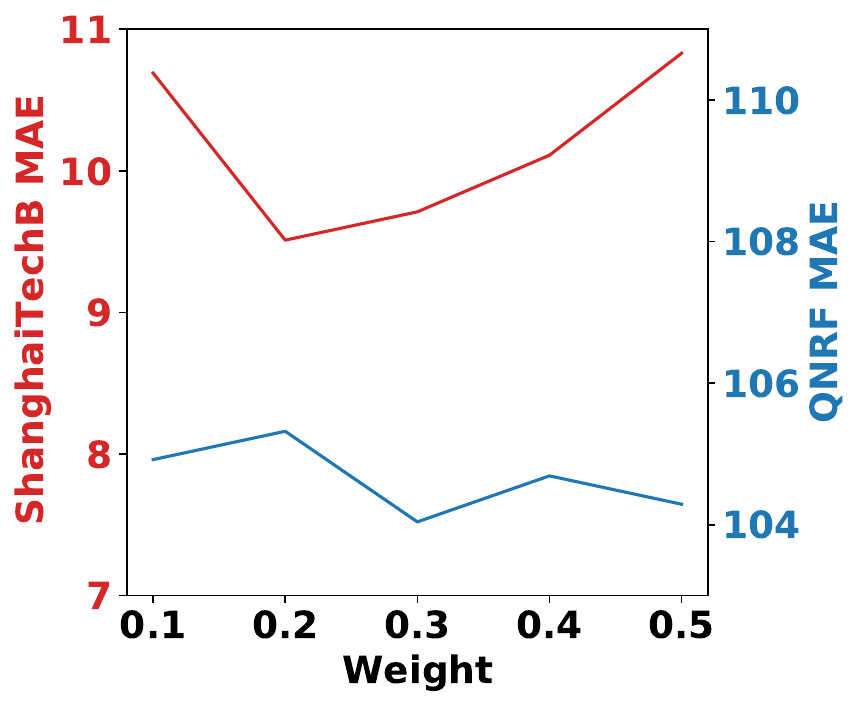}
  \end{subfigure}
  \begin{subfigure}{0.31\linewidth}
   \includegraphics[width=1\linewidth]{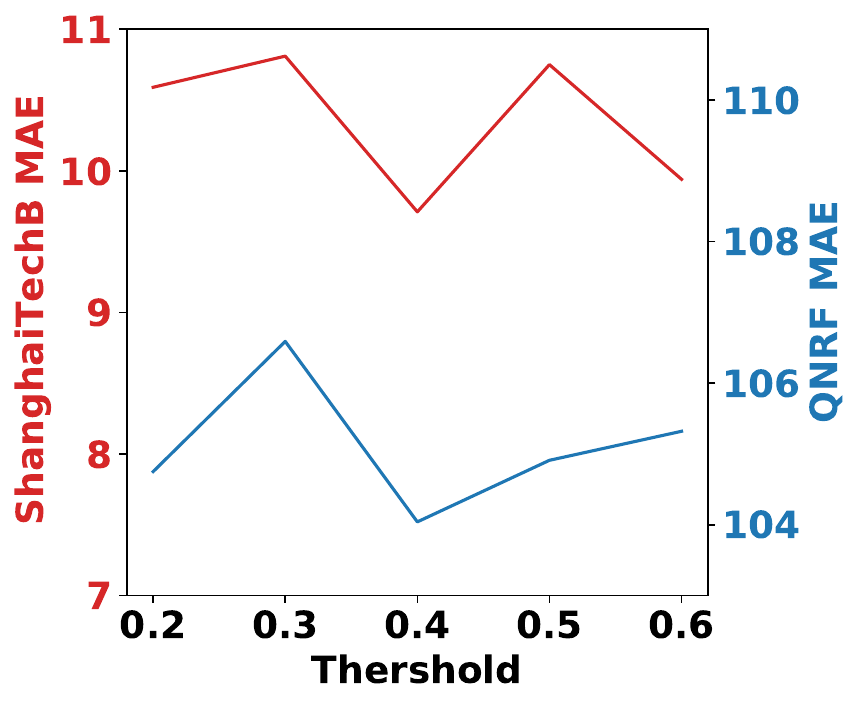}
  \end{subfigure}
  \caption{The hyperparameter ablation study results on ShanghaiTech B and UCF-QNRF.}
  \label{hyper}
\end{figure}

In Fig.~\ref{hyper}, we show additional ablation study results.

\subsection{Details of pseudo-labeling}
\label{sec:peudo-label}
Here we show the details of linearly increasing uncertainty threshold $u_t$ for choosing pseudo-labels:
\begin{equation*}
u_t =    startunc + \frac{endunc-startunc}{endep - startep} (t - startep)
\end{equation*}
where $u_t$ is the uncertainty threshold for choosing image patches, i.e., the image patches with uncertainty estimation higher than $u_t$ are blanked out, and $t$ is the current epoch number. The increase of $u_t$ begins at epoch $startep$ and ends at epoch $endep$. The uncertainty threshold increases from $startunc$ to $endunc$. By using this strategy, we can utilize high-quality model predictions at different training stages properly.

Since it takes several training iterations for multitask model to capture valid crowd and uncertainty information, we start leveraging unlabeled information from $10{th}$ epoch i.e. $startep = 10$. The model predictions on unlabeled images are error-prone, thus uncertainty threshold at the beginning $startunc$ is $0.1$. Besides, we have $endep = 130$ and $endunc = 0.6$.



\begin{table}[h]
\begin{center}
\begin{tabular}{ c|cc|cc } 
\hline
\multirow{2}{4em}{Method} & \multirow{2}{2em}{Type} & \multirow{2}{2em}{Ratio} & \multicolumn{2}{|c}{Part B}  \\
                           &        &              &MAE&RMSE\\
\hline
MT~\cite{tarvainen2017mean} & SSL& 5\% & 19.3 & 33.2  \\
L2R~\cite{liu2018leveraging} & SSL& 5\% & 20.3 & 27.6 \\
GP~\cite{sindagi2020learning} & SSL & 5\% & 15.7 & 27.9 \\
PA~\cite{xu2021crowd} & PAL &5\% & 16.50 & 25.28   \\
DAcount~\cite{lin2022semi} & SSL & 5\% & \underline{12.6} & \underline{22.8}  \\
ours &SSL & 5\% & \bf 11.03 & \bf 20.93  \\
\hline
MT~\cite{tarvainen2017mean} &SSL & 40\% & 15.9 & 25.7 \\
L2R~\cite{liu2018leveraging} & SSL & 40\% & 16.8 & 25.1 \\
DAcount~\cite{lin2022semi} &SSL & 40\% & \underline{9.6} & \underline{14.6} \\
Ours &SSL & 40\% & \bf 7.79 & \bf 12.70 \\
\hline

\end{tabular}
\caption{The ablation study results of labeled ratio on the ShanghaiTech part-B dataset.}
\label{rat_B}
\end{center}
\vspace{-.2in}
\end{table}

\subsection{Implementation details}
In practice, for the convenience of implementation, we use (1 - batch normalized ASM) as a surrogate to train the uncertainty branch, and the model confidence output will be used to filter out unreliable patches.

\end{document}